%% file: acl_latex.tex
\pdfoutput=1
\documentclass[11pt]{article}
\usepackage{amsmath,amssymb}
\usepackage{enumitem}
\usepackage{multirow}
\usepackage{stfloats}
\usepackage{listings}
\usepackage[]{acl}
\usepackage{amsmath}
\usepackage{times}
\usepackage{varwidth}
\usepackage{latexsym}
\usepackage{graphicx}
\usepackage[T1]{fontenc}

\usepackage[utf8]{inputenc}

\usepackage{microtype}

\usepackage{inconsolata}
\usepackage{authblk}
\usepackage[textsize=tiny]{todonotes}
\usepackage[listings]{tcolorbox}

\title{Revisiting Code Similarity Evaluation with Abstract Syntax Tree Edit Distance}

\author[1]{Yewei Song}
\author[1,2]{Cedric Lothritz}
\author[1]{Daniel Tang}
\author[1]{Tegawendé F. Bissyandé}
\author[1]{Jacques Klein}
\affil[1]{University of Luxembourg}
\affil[2]{Luxembourg Institute of Science and Technology}
\affil[1]{\textit {\{yewei.song, xunzhu.tang, tegawende.bissyande, jacques.klein\}@uni.lu}}
\affil[2]{\textit {\{cedric.lothritz\}@list.lu}}

\begin{document}
\maketitle
\begin{abstract}
This paper revisits recent code similarity evaluation metrics, particularly focusing on the application of Abstract Syntax Tree (AST) editing distance in diverse programming languages.
In particular, we explore the usefulness of these metrics and compare them to traditional sequence similarity metrics. 
Our experiments showcase the effectiveness of AST editing distance in capturing intricate code structures, revealing a high correlation with established metrics. Furthermore, we explore the strengths and weaknesses of AST editing distance and prompt-based GPT similarity scores in comparison to BLEU score, execution match, and Jaccard Similarity. 
We propose, optimize, and publish an adaptable metric that demonstrates effectiveness across all tested languages, representing an enhanced version of Tree Similarity of Edit Distance (TSED).

\end{abstract}

\section{Introduction and Related Work}
In the fields of natural language processing and software engineering, code generation tasks are gaining more and more attention. Assessing the quality of generated code is now critically important, but we still lack evaluation methods other than traditional statistical sequence evaluation methods. Widely used semantic evaluation metrics like BLEU score and Jaccard similarity rely on statistical characteristics, overlooking the intricate grammatical structures and logical relationships inherent in complex programming languages. 

However, recent developments in the NLP field paved the way for novel evaluation metrics which we explore in this study. For one, the staggering number of powerful large language models (LLMs) such as GPT-3.5/4~\cite{achiam2023gpt} revolutionized the NLP landscape and led to noteworthy advancements in the realm of code review and evaluation~\cite{wang2023software,tang2024collaborative}. Another recent study introduced the novel TSED metric and used it to evaluate text-to-SQL tasks~\cite{song2023enhancing}. For this study, we take advantage of these developments to (1) prompt the GPT-4 model to generate similarity scores for code, and (2) expand on the TSED metric.


We utilize these two different metrics (GPT and TSED) to evaluate the structural similarity of different programming languages and how they relate to execution matches. Furthermore, we address how these metrics are correlated to semantic similarity metrics like the BLEU score. Finally, we investigate some limitations of these metrics by delving into the impact of TSED's penalty weight of tree operations on evaluation accuracy and exploring the stability of outputs from the GPT LLMs.

As a result, we have these 3 contributions from this research: (a) we propose and publish a new tool for 48 programming languages\footnote{https://github.com/Etamin/TSED}, (b) we discuss 2 recent evaluation metrics and 2 traditional metrics and compare them via correlation coefficient, recall to execution match, (c) we discuss the unstable nature of GPT similarity scoring and the ways to optimize TSED.

\section{Approaches}
\subsection{TSED on Programming Languages}

\label{sec:tsed}
\input{latex/approaches}
\subsection{GPT Structure Similarity}
\label{sec:gpt}

Between 2020 and 2023, OpenAI introduced the GPT-3/3.5 and GPT-4 models, showcasing remarkable reasoning capabilities and achieving state-of-the-art performance across numerous tasks~\cite{brown2020language}. Our approach involves utilizing prompts to elicit the model's output regarding the structural similarity between two code segments, resulting in a score on a scale from 0 to 1. A score of 1 indicates identical structures, while 0 signifies complete dissimilarity. Despite its effectiveness, this metric operates as a black box, leaving us unaware of the specific calculations performed by GPT or whether it consistently employs the same metric. From various research papers, we've observed that these LLMs tend to produce more unstable results with each iteration~\cite{tian2023chatgpt,liu2023gpt}.
\begin{tcolorbox}
Given 2 \textbf{Java} code paragraphs, please generate a similarity score from 0 to 1 (to three decimal places), by grammar parsing structure. Answer with a format like [[0.777]].

=====Code 1=====

[\textbf{Java} code snippet 1]

=====Code 2=====

[\textbf{Java} code snippet 2]

=====End=====
\end{tcolorbox}
This prompt above is designed to calculate and return a similarity score between two Java code snippets based on their grammatical structure. The similarity score ranges from 0 to 1, with three decimal places of precision. A score of 1 indicates identical grammatical structures, while a score of 0 indicates completely different structures. The output format [[0.777]] facilitates easy extraction and post-processing of the score. 

\section{Research Questions and Targets}

\textbf{RQ1: Can TSED be used in more programming languages?} We investigate the adaptability of AST Edit Distance which is a generalized version of TSED, exploring its effectiveness in languages like Python and Java to assess its applicability for code similarity analysis.\\
\textbf{RQ2: How are TSED and GPT similarity correlated to semantic similarity and execution match?}  We assess the correlation between these different metrics to understand their respective contributions in evaluating code similarity across multiple programming languages. \\
\textbf{RQ3: What are the limits of these metrics?} We assess the stability of GPT-based similarity output and analyze how parameters, particularly operation weights (delete, insert, rename), influence TSED.

\section{Experiments}
\subsection{General Setup}
In this study, our primary objective is to apply the theoretical framework to a diverse range of programming languages. To achieve this, we aim to identify executable datasets and evaluate them using predefined metrics. The experimental setup comprises two key tasks: firstly, expanding the application of TSED and GPT similarity to additional programming languages, followed by exploring the correlation between these metrics. Subsequently, we seek to assess the stability of GPT scoring and examine the impact of various parameters on the TSED metric. This structured approach allows us to comprehensively investigate the adaptability, correlations, and stability of the chosen metrics across a spectrum of programming languages.


\subsection{Evaluation Metrics}
\begin{itemize}[leftmargin=*,labelsep=0.1cm]
    \item \textbf{BLEU Score} is calculated as the geometric mean of the modified precision scores for various n-gram lengths, providing a concise and standardized similarity measurement between the generated and reference text~\cite{10.3115/1073083.1073135}.
    \item \textbf{Jaccard Similarity} is a measure of similarity between two sets and is calculated by dividing the size of the intersection of the sets by the size of their union, offering a quantitative assessment of the degree of overlap between the sets' elements.
    \item \textbf{Execution Match} Execution Match pertains to the consistency in execution outcomes between generated code and its corresponding ground truth, evaluating the equivalence in practical functionality. 1 in Execution match means they have the same execution results, and 0 means different.
    \item \textbf{GPT Similarity} mentioned in the Section \ref{sec:gpt}
    \item \textbf{TSED} mentioned in the Section \ref{sec:tsed}.
\end{itemize}

\subsection{Datasets}
Although the execution match metric is infrequently employed in programming code-related datasets, its prominence has increased in recent years. Our comparative analysis involved assessing datasets from various papers, considering factors such as dataset sizes, programming languages, and executables. As highlighted in Table \ref{tab:benchmarks}, the \textbf{MBXP} dataset encompasses 13 different languages, serving as a function-level benchmark that effectively evaluates programming paragraphs. However, the MBXP dataset includes ground-truth solutions for only 7 languages, with C\# omitted due to compilation issues. Additionally, we consider the \textbf{CoderEval} dataset to facilitate a comparison between Python and Java code generation, leveraging its longer test samples, results are in the appendix.

\begin{table}[htbp]
\caption{Widely-used code generation benchmarks, selected from GitHub}
\label{tab:benchmarks}
\centering
\resizebox{\columnwidth}{!}{%
\begin{tabular}{lllll}
\hline
Benchmark & Language       & Samples   & Executeable \\ \hline
CoNaLA\cite{yin2018learning}    & Python        & 500       & No          \\
Concode\cite{iyer2018mapping}   & Java                  & 2000      & No          \\
\textbf{MBXP}\cite{athiwaratkun2022multi}      & Multilingual          & 974       & Yes         \\
\textbf{InterCode}\cite{yang2023intercode} & Bash, SQL             & 200, 1034 & Yes         \\
\textbf{CoderEval}\cite{yu2024codereval} & Python, Java          & 230       & Yes         \\
RepoEval\cite{liao2023context}  & Python            & 383       & No          \\ \hline
\end{tabular}%
}
\vspace{-1.0em}
\end{table}

In the Bash-Shell scenarios, we reproduce results and conduct a comparative analysis using the \textbf{InterCode} dataset. Notably, we identify the SPIDER dataset within InterCode and establish it as a baseline. \textbf{SPIDER}, previously evaluated in comparison to the TSED paper, is a substantial human-labeled dataset for the text-to-SQL task. This dataset encompasses databases with intricate join solutions across diverse domains~\cite{yu2018spider}.

\section{Results}
\subsection{Similarity Results}
\begin{table}[htbp]
\caption{Evaluation Metrics comparison for 6 languages on MBXP dataset, prediction generated by GPT-3.5-Turbo model, ground truth from dataset}
\label{tab:mbxp-gpt}
\centering
\resizebox{\columnwidth}{!}{%
\begin{tabular}{lrrrrr}
\hline
Languages & \multicolumn{1}{l}{TSED} & \multicolumn{1}{l}{BLEU} & \multicolumn{1}{l}{Jaccard Sim} & \multicolumn{1}{l}{GPT-4} & Execution \\ \hline
Java       & 0.3746 & 0.2041 & 0.2733 & 0.8143 & 0.6550 \\
Python     & 0.1888 & 0.0843 & 0.2000 & 0.6751 & 0.6842 \\
JavaScript & 0.2037 & 0.0846 & 0.2037 & 0.6763 & 0.6811 \\
Typescript & 0.1360 & 0.0637 & 0.1397 & 0.5313 & 0.6642 \\
Ruby       & 0.1727 & 0.0438 & 0.1810 & 0.7067 & 0.6428 \\
Kotlin     & 0.3412 & 0.1847 & 0.3109 & 0.7073 & 0.5569 \\ \hline
\end{tabular}%
}
\vspace{-1em}
\end{table}

\begin{table*}[htb]
\caption{Execution Match F1 score \& Accuracy for each thresholding metrics}
\label{tab:recall}
\centering
\resizebox{1\textwidth}{!}{%
\begin{tabular}{l|rrr|rrr|rrr|rrr}
\hline
  Languages &
  \multicolumn{3}{c}{TSED} &
  \multicolumn{3}{c}{GPT} &
  \multicolumn{3}{c}{BLEU} &
  \multicolumn{3}{c}{Jaccard} \\ \cline{2-13} 
 &
  \multicolumn{1}{l}{Threshold} &
  \multicolumn{1}{l}{F1} &
  \multicolumn{1}{l}{Acc} &
  \multicolumn{1}{l}{Threshold} &
  \multicolumn{1}{l}{F1} &
  \multicolumn{1}{l}{Acc} &
  \multicolumn{1}{l}{Threshold} &
  \multicolumn{1}{l}{F1} &
  \multicolumn{1}{l}{Acc} &
  \multicolumn{1}{l}{Threshold} &
  \multicolumn{1}{l}{F1} &
  \multicolumn{1}{l}{Acc} \\ \hline
Python &
  0.23 &
  0.5650 &
  0.6057 &
  0.83 &
  \underline{0.6403} &
  \textbf{0.6735} &
  0.07 &
  0.5719 &
  0.6150 &
  0.19 &
  0.5907 &
  0.6253 \\
Java &
  0.10 &
  0.5108 &
  \textbf{0.6499} &
  0.56 &
  \underline{0.5693} &
  0.6396 &
  0.03 &
  0.5184 &
  0.5755 &
  0.16 &
  0.5612 &
  0.6018 \\
JavaScript &
  0.12 &
  0.5494 &
  0.6002 &
  0.69 &
  \underline{0.5924} &
  \textbf{0.6205} &
  0.02 &
  0.4964 &
  0.5267 &
  0.12 &
  0.5245 &
  0.5885 \\
Typescript &
  0.07 &
  0.5367 &
  \textbf{0.5822} &
  0.51 &
  \underline{0.5521} &
  0.5708 &
  0.01 &
  0.4987 &
  0.5553 &
  0.08 &
  0.5284 &
  0.5708 \\
Ruby &
  0.13 &
  0.5045 &
  0.5306 &
  0.54 &
  \underline{0.6051} &
  \textbf{0.6811} &
  0.01 &
  0.4375 &
  0.4490 &
  0.12 &
  0.5142 &
  0.5612 \\
Kotlin &
  0.28 &
  \underline{0.6834} &
  \textbf{0.6823} &
  0.8 &
  0.6681 &
  0.6721 &
  0.1 &
  0.6441 &
  0.6457 &
  0.22 &
  0.6387 &
  0.6533 \\ \hline
\end{tabular}%
}
\end{table*}
As we analyze the results presented in Table \ref{tab:mbxp-gpt}, our experiment demonstrates the effective performance of TSED and GPT similarity in evaluating the MBXP dataset across all 6 programming languages. No instances of parsing or scoring generation failures were observed, confirming the robustness of these metrics across languages.

\begin{tcolorbox}[leftrule=0mm,rightrule=0mm,toprule=0mm,bottomrule=0mm,left=0pt,right=0pt,top=0pt,bottom=0pt,title={RQ1: Can TSED be used in more programming languages?}]
    \textbf{Answer: }The exploration of TSED's adaptability beyond SQL shows promise, especially in languages like Java and Kotlin, indicating its potential for code analysis. TSED proves effective in programming languages with functional parsers, allowing for structural similarity calculation.
\end{tcolorbox}
\vspace{-1em}
\begin{figure}[htbp]
    \centering
    \includegraphics[width=0.43\textwidth]{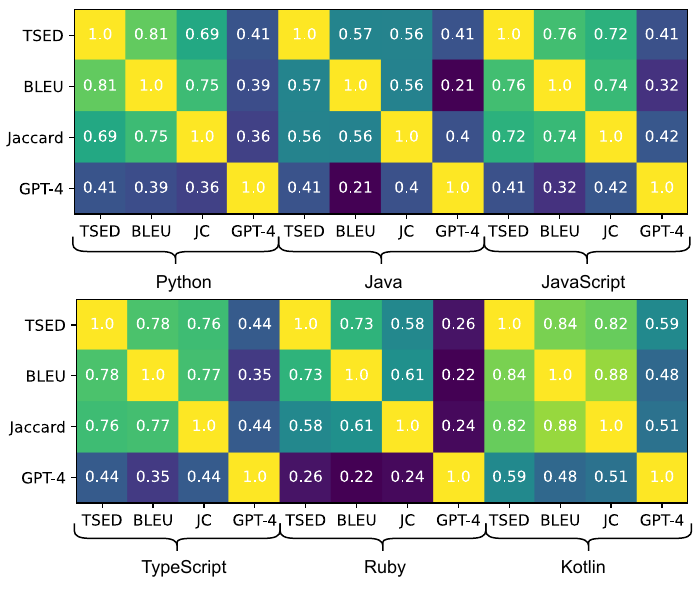}
    \caption{MBXP dataset, Pearson Correlation Heatmap between evaluation-metrics on GPT-3.5}
    \label{fig:heatmapce}
    \vspace{-1.0em}
\end{figure}

\begin{figure}[htbp]
    \centering
    \hspace{-1cm}\includegraphics[width=0.45\textwidth]{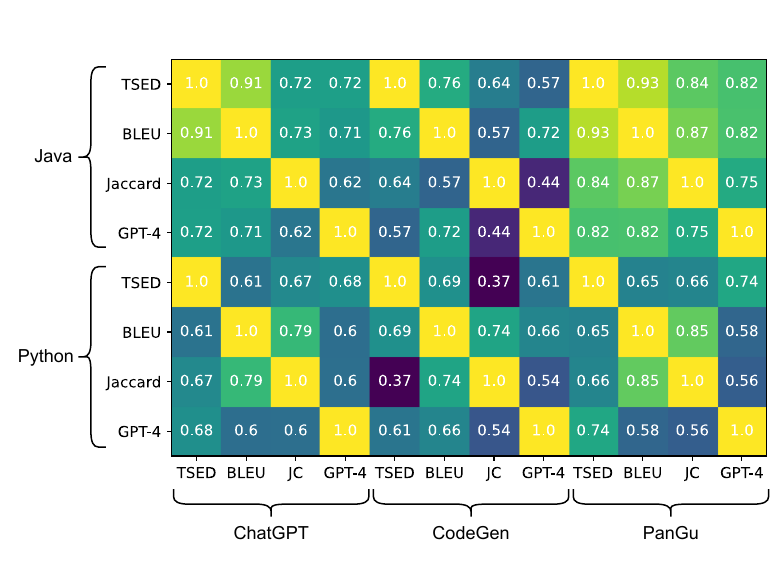}
    \caption{CoderEval Pearson Correlation Heatmap between evaluation-metrics/models/languages}
    \label{fig:heatmapcodereval}
    \vspace{-1.5em}
\end{figure}

Moreover, TSED shows a commendable correlation ranging from 0.6 to 0.8 with BLEU score and Jaccard similarity, as illustrated in Figure \ref{fig:heatmapce}. Additionally, TSED exhibits a strong correlation with GPT similarity, especially in Java and Python during the CoderEval test, as depicted in Figure \ref{fig:heatmapcodereval}, underscoring its sensitivity to code structure.
We employ thresholding to establish a prediction-to-execution match. If the metric value exceeds the threshold $T$, we assign the prediction as 1; otherwise, it is set to 0. The optimal threshold values are determined through enumeration to achieve the best match results. Based on their F1/Accuracy match to the Execution match, both TSED and GPT similarity exhibit higher accuracy compared to semantic metrics in Table \ref{tab:recall}. Notably, GPT similarity demonstrates a slightly superior F1 score and TSED gives good results on accuracy.

\begin{tcolorbox}[leftrule=0mm,rightrule=0mm,toprule=0mm,bottomrule=0mm,left=0pt,right=0pt,top=0pt,bottom=0pt,title={RQ2: How are TSED and GPT similarity correlated to semantic similarity and execution match?}]
    \textbf{Answer: }Our evaluation of  TSED metrics, GPT-based similarity, and other semantic evaluation metrics revealed consistently high Pearson correlations between TSED, GPT Score, BLEU Score, and Jaccard Similarity. TSED exhibited notable accuracy in matching with Execution-Match, while GPT score demonstrated the highest F1 score, highlighting their respective strengths in capturing structural and semantic nuances in code across various programming languages.
\end{tcolorbox}

\subsection{Stability of GPT Scoring}
To understand how unstable GPT scoring is, we execute the GPT-4 Similarity scoring five times on identical prediction sets, we establish the initial result as a baseline to assess differences through statistical indicators such as Mean Squared Error (MSE) or Mean Absolute Error (MAE) in comparison to the first scoring. Table \ref{tab:jitter} demonstrates that GPT scoring exhibits limited stability in the context of code similarity evaluation. 
\begin{table}[htpb]
\caption{Unstable nature of GPT-4 scoring output}
\label{tab:jitter}
\centering
\resizebox{0.9\columnwidth}{!}{%
\begin{tabular}{lllll}
\hline
Metrics             & 1st    & 2nd    & 3rd    & 4th    \\ \hline
Mean Squared Error  & 0.0581 & 0.0583 & 0.0527 & 0.0628 \\
Mean Absolute Error & 0.1902 & 0.1940 & 0.1825 & 0.1996 \\ \hline
\end{tabular}%
}
\vspace{-1em}
\end{table}

\subsection{Parameter optimization of TSED}
\begin{figure}[htbp]
    \centering
    \includegraphics[width=0.45\textwidth]{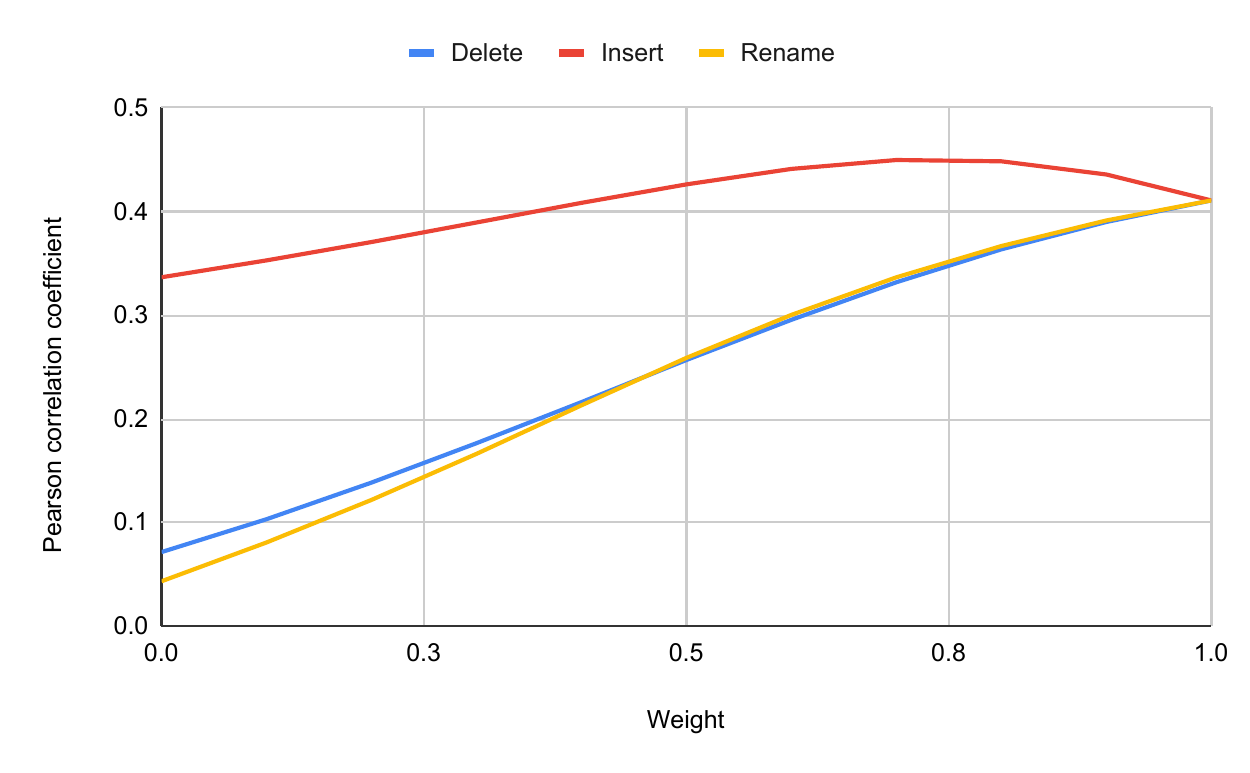}
    \caption{Change each of penalty weight influence correlation to GPT structure similarity score}
    \label{fig:weight}
\end{figure}

We can configure the penalty weight of 3 operations in tree distance computing: \textbf{Delete}, \textbf{Insert}, and \textbf{Rename}. Figure \ref{fig:weight} which is from a test for the MBXP/Java dataset shows is `Insert' has a sweet spot of 0.8. 'Delete' and 'Rename' operations just keep them in 1.0  penalty weight as the best choice. But we need to keep in mind it can be different in other programming languages.
\begin{tcolorbox}[leftrule=0mm,rightrule=0mm,toprule=0mm,bottomrule=0mm,left=0pt,right=0pt,top=0pt,bottom=0pt,title={RQ3: What are the limits of these metrics?}]
    \textbf{Answer: } Penalty weight parameters play influential roles in the TSED metric. Besides, GPT-based similarity metrics offer higher performance at the cost of more money, leading to a bit of unstable output. This underscores the need to carefully balance performance and stability considerations in code similarity assessment across various programming languages.
\end{tcolorbox}
\subsection{Efficiency}

\begin{table*}[hbp]
\caption{Average execution time(ms) of metrics and programming languages}
\label{tab:efficiency}
\centering
\resizebox{1.6\columnwidth}{!}{%
\begin{tabular}{llllllll}
\hline
\textbf{} & \textbf{Python} & \textbf{Java} & \textbf{JavaScript} & \textbf{TypeScript} & \textbf{C\#} & \textbf{Ruby} & \textbf{Kotlin} \\ \hline
TSED      & 0.0227 & 0.0645 & 0.0315 & 0.0697 & 0.0373 & 0.0092 & 0.0307 \\
BLEU      & 0.0075 & 0.0113 & 0.0155 & 0.0163 & 0.0160 & 0.0116 & 0.0144 \\
Jaccard   & 1.6e-5 & 2.9e-5 & 1.9e-5 & 2.4e-5 & 2.7e-5 & 1.5e-5 & 1.8e-5 \\
GPT3.5ß Score & 1304   & 1860   & 1231   & 1339   & 1470   & 1044   & 1681   \\ \hline
\end{tabular}%
}
\end{table*}
The table \ref{tab:efficiency} illustrates the computational time (in ms) required by each programming language tested, including TSED, BLEU score, Jaccard similarity, and GPT 3.5 Score. Our findings indicate that the performance of TSED is comparable to the BLEU score, with significantly lower computational time compared to GPT-3.5. This suggests that TSED is indeed efficient enough to be applied at scale.
\section{Conclusion}
In this paper, we applied TSED to more programming languages, compared GPT similarity and TSED to semantic metrics, and checked representation to execution match. Then we discuss limitations about the stability of GPT scoring and the penalty parameters of TSED. 
\section*{Limitations}
While our study provides valuable insights into code similarity assessment using TSED and GPT-based metrics, it is essential to acknowledge certain limitations. Firstly, the generalizability of our findings may be influenced by the specific datasets and programming languages employed in our analysis. Additionally, the stability of GPT-based similarity metrics, as highlighted in our results, poses a limitation in terms of consistent and reliable code assessments. Furthermore, variations in the interpretation and definition of similarity metrics across different studies may introduce inherent biases. Lastly, the effectiveness of TSED metrics may be contingent upon the quality of the employed parsers and the fine-tuning of penalty parameters. These limitations underscore the need for caution when extrapolating our results to diverse contexts and emphasize the necessity for further research to address these challenges.
\section*{Ethics Statement}
Our research adheres to ethical standards, prioritizing integrity and respect for all involved parties. We ensured data privacy, obtained informed consent where applicable, and maintained transparency in our methodologies. The study was conducted with the utmost consideration for ethical guidelines and the welfare of participants, upholding the principles of fairness, accountability, and academic integrity throughout the research process.

\section*{Acknowledgment}
This research was funded in whole, or in part, by the Luxembourg National Research Fund (FNR), grant references NCER22/IS/16570468/NCER-FT and  BRIDGES2021/IS/16229163/LuxemBERT. 
We extend our heartfelt appreciation to our collaborator, BGL BNP PARIBAS, for their invaluable support and special thanks to Saad Ezzini from Lancaster University for his advisory contributions.

\bibliography{anthology, custom}

\appendix
\label{sec:appendix}
\input{latex/appendix}

\end{document}

%% file: latex/approaches.tex
\begin{figure*}[htb]
\centering
\includegraphics[width=1.8\columnwidth]{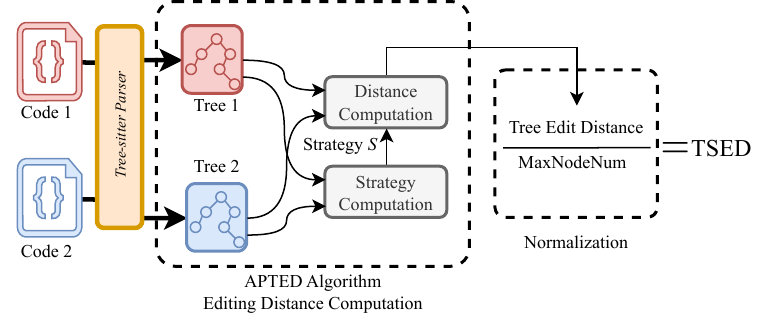}
\caption{Pipeline of TSED Code Evaluation Metric}
\label{fig:tsed}
\vspace{-1.0em}
\end{figure*}

Applying the TSED evaluation method, initially designed for SQL analysis, we have undergone modifications to extend its applicability to various programming languages. The fundamental TSED approach, illustrated in Figure \ref{fig:tsed}, encompasses AST parsing, AST Editing Distance Calculation, and normalization, closely resembling the methodology outlined in the original paper. However, we have made modifications to both the AST parsing and normalization.

\textbf{Code Parsing:} Parsing in the domain of programming languages involves parsing raw code text into its associated AST. This parsing underscores the complexity of interpreting various programming constructs and converting them into a structured grammar tree representation. 



We use tree-sitter\footnote{https://tree-sitter.github.io/tree-sitter/} as our AST parser which is based on GLR(generalized left-to-right rightmost), a powerful parsing algorithm commonly found in the literature~\cite{latif2023comparison,tomita1991generalized,clem2021static}.

\textbf{Tree Distance Computation:} For calculating tree edit distance as $\Delta$, we utilize the same function as outlined in the TSED paper, which is APTED(All Path Tree Edit Distance) algorithm~\cite{pawlik2015efficient, pawlik2016tree}. 
Considering $G_1$ as predicted code's AST and $G_2$ as AST from ground-truth: \vspace{-0.5em}

\begin{equation}
\Delta(G_1, G_2) = \min_{ops} \sum_{i=1}^{n} w(op_i)
\end{equation}

Here, \( ops \) is a sequence of edit operations transforming \( G_1 \) into \( G_2 \), with \( w(op_i) \) as the cost for the \( i^{th} \) operation.

\textbf{Normalization:} Normalization of tree edit distances accounts for the complexity of the code by considering the maximum number of nodes between two trees, and we add a ramp function to avoid some extreme situations: 
\begin{equation}\hspace{-0.18cm}
TSED = \max\{1-\frac{\delta}{MaxNodes(G_1, G_2)},0\}
\end{equation}

This provides a metric for structural similarity comparison of programming code, enabling a nuanced analysis beyond mere syntactic comparison.

%% file: latex/appendix.tex
\newpage
\section{Additional Experiment Details}
\subsection{Parser Comparison}
\begin{figure}[htbp]
    \centering
    \includegraphics[width=0.45\textwidth]{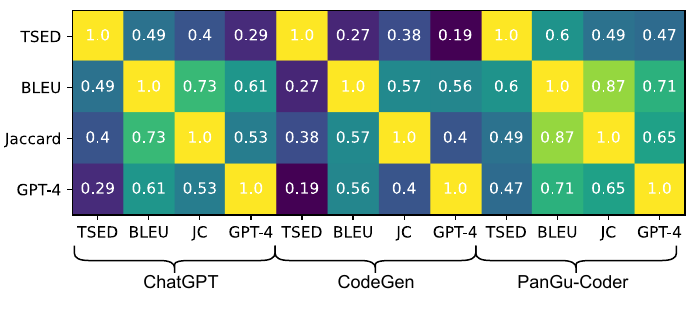}
    \caption{CoderEval Java Pearson Correlation Heatmap between evaluation-metrics/models/languages on TSED with ANTLR parser}
    \label{fig:heatmapantlr}
\end{figure}

The ANTLR\footnote{https://www.antlr.org/} (ANother Tool for Language Recognition) tool, serving as a distinct AST parser compared to tree-sitter, demonstrated notable differences. Following our evaluation using identical settings for TSED metrics, as Figure \ref{fig:heatmapantlr} shows, it became evident that the correlation with other metrics was inferior to the original solutions. This experiment underscores the crucial role of parser performance in the computation procedure, highlighting the significance of selecting an appropriate parser for accurate and reliable code similarity assessments.

\subsection{Other experiment results}

Due to space constraints, a subset of experimental data is provided in the appendix. A comprehensive evaluation of CoderEval and InterCoder is detailed in Table \ref{tab:codereval}, while specific original sample data from the MBXP dataset is presented in Table \ref{tab:mbxp}.

CoderEval, designed for class-level code generation tasks, proves to be a challenging test. Utilizing Pass@10 data as a test sample, TSED demonstrates a robust correlation with semantic indicators in both Java and Python languages. Additionally, a noteworthy correlation is observed between TSED and GPT Similarity.

\begin{figure}[htbp]
    \centering
    \includegraphics[width=0.45\textwidth]{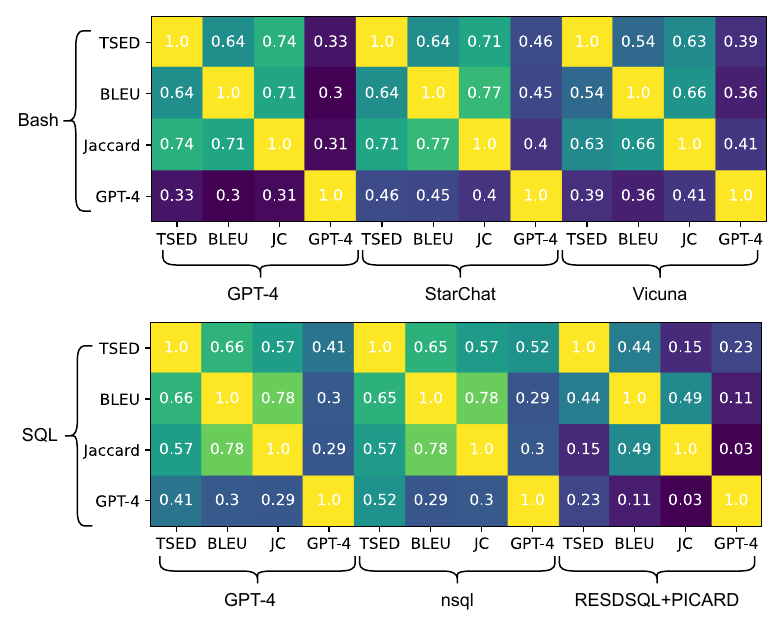}
    \caption{InterCode/SPIDER Pearson Correlation Heatmap between evaluation-metrics/models/languages}
    \label{fig:heatmapinter}
\end{figure}

In the case of InterCoder, we confirm that TSED calculations extend to Bash scripts. Also, the correlation in Figure \ref{fig:heatmapinter} between TSED to semantic metrics is acceptable, the GPT score doesn't have a good correlation to others. We also replicate the performance of the SPIDER dataset, noting differences from the original paper but not to a significant extent.

Despite the notably low semantic similarity between the MBXP built-in samples and the ground truth, a relatively high execution match is observed. We acknowledge this disparity and plan to address it through optimization in future research endeavors.

\begin{table}[htbp]
\caption{4 Evaluation Metrics compared to Ground Truth on CoderEval(Java\&Python) / InterCode(Bash) / SPIDER(SQL)}
\label{tab:codereval}
\centering
\resizebox{\columnwidth}{!}{%
\begin{tabular}{llccccc}
\hline

\hline

\hline

\hline
Languages & Model & \multicolumn{1}{l}{TSED} & \multicolumn{1}{l}{BLEU} & \multicolumn{1}{l}{Jaccard Sim} & \multicolumn{1}{l}{GPT-4} & Execution \\ \hline
Java   & ChatGPT & 0.4971 & 0.3655 & 0.3384 & 0.7392 & 0.3539 \\
                        & CodeGen & 0.3616 & 0.2871 & 0.2506 & 0.6603 & 0.1391 \\
                        & PanGu   & 0.5029 & 0.3722 & 0.3849 & 0.6778 & 0.2543 \\ \hline
Python & ChatGPT & 0.2840 & 0.1285 & 0.1763 & 0.5883 & 0.2104 \\
                        & CodeGen & 0.2703 & 0.1778 & 0.1821 & 0.5604 & 0.0948 \\
                        & PanGu   & 0.2829 & 0.0868 & 0.1567 & 0.5086 & 0.1183 \\ \hline
                    \multirow{3}{*}{Shell} & GPT-4    & 0.5853 & 0.2816 & 0.3567 & 0.8511 & 0.4851 \\
                       & starchat & 0.4065 & 0.1594 & 0.2081 & 0.6740 & 0.2374 \\
                       & vicuna   & 0.4755 & 0.1621 & 0.2295 & 0.7164 & 0.2451 \\ \hline
\multirow{3}{*}{SQL}   & ChatGPT-3.5    & 0.6824 & 0.3304 & 0.3710 & 0.9461 & 0.6482 \\
                       & nsql-6B  & 0.8022 & 0.4493 & 0.4356 & 0.9265 & 0.5483 \\
                       & RESDSQL  & 0.7422 & 0.2084 & 0.1868 & 0.9629 & 0.7756 \\ \hline

                       \hline

                       \hline

                       \hline
\end{tabular}%
}
\end{table}

\begin{table}[htbp]
\caption{4 Evaluation Metrics compare to Ground Truth on 7 languages MBXP Dataset Samples}
\label{tab:mbxp}
\resizebox{\columnwidth}{!}{%
\begin{tabular}{lccccc}
\hline

\hline

\hline

\hline
Languages & \multicolumn{1}{l}{TSED} & \multicolumn{1}{l}{BLEU} & \multicolumn{1}{l}{Jaccard Sim} & \multicolumn{1}{l}{GPT-4} & Execution \\ \hline
Java       & 0.2218 & 0.1046 & 0.1960 & 0.4248 & 0.853 \\
Python     & 0.1550 & 0.0255 & 0.1222 & 0.3396 & 0.822 \\
JavaScript & 0.1870 & 0.0573 & 0.1685 & 0.4005 & 0.786 \\
Typescript & 0.1186 & 0.0288 & 0.1260 & 0.4247 & 0.872 \\
Ruby       & 0.2073 & 0.0235 & 0.1796 & 0.4830 & 0.589 \\
Kotlin     & 0.1720 & 0.0336 & 0.1877 & 0.3976 & 0.637 \\ \hline

\hline

\hline

\hline
\end{tabular}%
}
\end{table}
\section{Case Studies}
\lstset{
  basicstyle=\ttfamily,
  columns=fullflexible,
  frame=single,
  breaklines=true,
  postbreak=\mbox{\textcolor{red}{$\hookrightarrow$}\space},
}
\subsection{A. Low BLEU, but high TSED}
\begin{lstlisting}[linewidth=\columnwidth,language=Python]
### Code Paragraph 1
int result = 0;
        for(int i = 0; i < n; i++) {
            result = n * (7 * n - 5) / 2;        
}
        return result;
    }
}
### Code Paragraph 2
int jacobsthalNumber = 1;
        for(int i = 2; i <= n; i++){
            jacobsthalNumber = jacobsthalNumber + (n - i) * (i - 1);
        }
        return jacobsthalNumber;
    }
}
\end{lstlisting}

In the provided code snippets, both segments involve loops for performing calculations, which contributes to their high structural similarity. However, the semantic similarity is relatively low due to the significant disparity in variable names, which occupy a considerable portion of the tokens. Despite the differences in semantics, the BLEU score, a metric commonly used for evaluating text similarity, yields a score of 0.359, indicating some level of similarity. In contrast, the Tree Similarity of Edit Distance (TSED) metric, which accounts for structural differences, produces a higher score of 0.8, highlighting the effectiveness of TSED in capturing structural similarities even when semantic differences exist.

\subsection{BLEU and TSED similar}
\begin{lstlisting}[language=Python]
### Code Paragraph 1
def max_of_two(a, b):
    if a > b:
        return a
    else:
        return b
### Code Paragraph 2
def max_of_two(a, b):
    return max(a, b)
\end{lstlisting}
Both the BLEU score (0.408) and the TSED (0.444) score suggest that while the two code snippets achieve somehow similar functionality, they do so using different structural approaches.

\textbf{Token Similarity}: Despite structural differences, the two snippets share many common tokens, leading to a moderate BLEU score. This reflects their similar semantic intent and consistent use of basic programming constructs.

\textbf{Structural Difference}: The TSED score highlights the structural changes. The first snippet uses explicit control flow statements (if-else), while the second snippet uses a built-in function (max). This difference is significant enough to lower the structure similarity score but not so drastic as to suggest entirely different functionalities.

Overall, these scores indicate that the snippets are functionally similar (performing the same task) with some commonality in token usage but exhibit distinct structural approaches in their implementations. This case illustrates how different coding styles and structures can be used to achieve the same outcome in programming.
\subsection{Let ChatGPT explain in detail}
\begin{tcolorbox}
    \textbf{Prompt: }Given 2 Java code paragraphs, please generate a similarity score from 0 to 1 (to three decimal places), by grammar parsing structure. Answer with a format like [[0.777]]. And give us 3 dimensions of similarity in integer score 1~5, which is “functional”, “tree structure”, and “semantical”; Then give the reason of the score you think if it’s high or low.
\end{tcolorbox}

\begin{lstlisting}[linewidth=\columnwidth,language=Python]
int result = 0;
        for(int i = 0; i < n; i++) {
            result = n * (7 * n - 5) / 2;        
}
        return result;
    }
}
int jacobsthalNumber = 1;
        for(int i = 2; i <= n; i++){
            jacobsthalNumber = jacobsthalNumber + (n - i) * (i - 1);
        }
        return jacobsthalNumber;
    }
}
\end{lstlisting}
Here is the similarity score and the three dimensions of similarity for the given code snippets:
[[0.875]]
Functional Similarity: 4
Tree Structure Similarity: 5
Semantical Similarity: 3
Explanation:
\begin{itemize}
    \item Functional Similarity (4): Both code snippets perform calculations within a loop structure, albeit with different mathematical operations. They both iterate over the variable i to calculate a result based on the value of n.
    \item Tree Structure Similarity (5): The overall structure of both code snippets is highly similar, with both containing a loop structure followed by a calculation and a return statement.
    \item Semantical Similarity (3): While the overall structure is similar, the semantics differ significantly. The variable names (result vs. jacobsthalNumber) and the mathematical operations performed within the loop differ, leading to a moderate level of semantic dissimilarity.
\end{itemize}